\documentclass[10pt,twocolumn,letterpaper]{article}

\usepackage{icb}
\usepackage{times}
\usepackage{epsfig}
\usepackage{graphicx}
\usepackage{amsmath}
\usepackage{amssymb}

\DeclareMathOperator{\arctantwo}{arctan2}

\icbfinalcopy 

\ificbfinal\fi
\begin{document}

\title{Comparison of fingerprint authentication algorithms for small imaging sensors}

\author{Mathilde Bourjot\\
\and
Regis Perrier\\
CEA Leti - Grenoble
Systems Department\\
{\tt\small firstname.lastname@cea.fr}
\and
Jean François Mainguet
}

\maketitle
\thispagestyle{empty}

\begin{abstract}
The demand for biometric systems has been increasing with the growth of the smartphone market. Biometric devices allow the user to authenticate easily while securing its private data without the need to remember any access code. Amongst them, fingerprint sensors are the most widespread because they seem to provide a good balance between reliability, cost and ease of use. According to smartphone manufacturers, the security level is guaranteed to be high. However, the size of those sensors, which is only a few millimeters squared, prevents the use of minutiae algorithms. To the best of our knowledge, very few studies shed light onto this problem, though many pattern recognition algorithms already exist as well as commercial solutions which are supposedly robust. In this article we try to provide insights on how to tackle this problem by analyzing the performance of three algorithms dedicated to pattern recognition.
\end{abstract}

\section{Introduction}
\begin{figure}
\center
\includegraphics[scale=0.4]{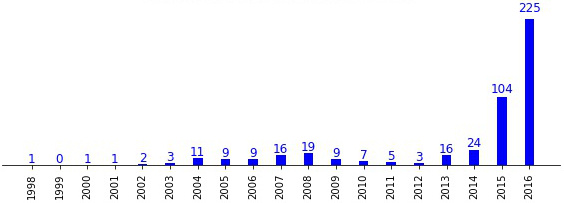}
\caption{New fingerprint enabled smartphone models per year (source: mainguet.org).}
\label{fig_market}
\vspace{-0.3cm}
\end{figure}

The importance of fingerprint for human identification does not need to be demonstrated; governments have been using them since several dozens of years and fingerprint systems were first integrated in personal digital assistants (PDA) and laptops in the early 2000s. At that time, a few millions of sensors were sold each year. Nowadays, since the launch of the iPhone 5s with its Touch ID system in 2013, sales have reached 1 billion sensors per year. The market growth is well illustrated in figure \ref{fig_market}.

Optical sensors with a large acquisition area of a few centimeters squared are the standard for governmental applications. This contrasts with the requirement of smartphone manufacturers for which sensors need to be as small as a few millimeters squared for reasons such as cost and electronic integration.\footnote{The iPhone 5s has a 4.5x4.5mm${}^2$ sensor.} 
This small size happens to be problematic for recognition tasks and raises the question of its reliability.

Minutiae algorithms have been used as an international standard for governmental use. In the case of a swipe fingerprint sensor, it has been demonstrated that a minimum overlap of 7mm along the height of a pair of images is mandatory to perform a match with those algorithms \cite{Mainguet04}. Recent work \cite{Madrid17} also confirmed that for a fixed False Acceptance Rate (FAR) of 1 for 10000, the bigger the sensor acquisition surface is, the better are the False Rejection Rates (FRR): from 1\% for a large sensor, it degrades to 20\% for a 8x8mm${}^2$ sensor.
\begin{figure}
\center
\includegraphics[scale=1.8]{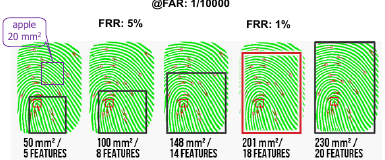}
\caption{Recognition rates with respect to sensor size for minutiae algorithms \cite{Mainguet04}.}
\vspace{-0.3cm}
\end{figure}

Those studies focused on matching algorithms which are based on minutiae, they correspond to second level features in a fingerprint well known for their persistence and uniqueness amongst individuals. The first level features characterizes the overall shape of the finger at a macroscopic level, and the third considers ridges variations and sweat pores. Given recognition rates promoted by smartphone makers,\footnote{FAR of 1/50000 with no FRR announced for the iPhone 5s.} it seems likely that their authentication algorithms use a mixture of all of those features. In this article, we compare three standard algorithms from the computer vision literature which could be suited for the application of fingerprint matching using small imaging sensors: one belonging to the so called direct methods and two known as feature based methods. If the literature is rich of methods to preprocess raw images and extract minutiae, up to our knowledge, only few have promoted alternative methods for the case of small sensors. Noticeable works are \cite{Lee99,Jain00}: they use Gabor filter banks to construct a small vector of features called fingercode and achieve recognition rates close to minutiae based algorithms. However, they consider fingerprint images of standard size. One of the first mention to the smartphone case appeared recently in \cite{Karlsson16}, where the authors have performed an exhaustive analysis of feature based methods and for which SIFT outperforms the others; this will contradict our analysis. 

This article provides a baseline for authentication performance that could be reached using small fingerprint sensors and standard computer vision algorithms. It also gives advices on how to improve those results. 

\section{Biometric vocabulary}
Biometric systems have a two stage process \cite{Jain04}.
\begin{itemize}
\item The first one is the enrollment phase in which physiological features of the user are recorded; this is an image of the fingerprint in our case. Some process reduces the original image into features that are stored in memory.
\item The second step is the recognition or authentication which consists in acquiring a candidate image and compare it to the one already stored during the enrollment phase. A matching algorithm will then compute a similarity score and will decide according to some predefined threshold if the candidate image is sufficiently close to the enrolled one.
\end{itemize}

In the specific case of small fingerprint sensors, the user will be asked to repeatedly submit his finger to the sensor during the enrollment so that a sufficient large area of it is captured. This will increase chances of recognizing a smaller portion of the finger thereafter. Quite logically, it is mandatory to display the same finger portion during enrollment that will be used for authentication afterward.

Two rates are usually defined:
\begin{itemize}
\item FAR (False Acceptance Rate): this rate is related to security issues. It has to be around 1 over 10000 for a mass market product, as opposed to the police requirements that are of 1 over a million in order to look for people in big databases.
\item FRR (False Rejection Rate): this one is experienced by the user and is related to a commodity issue. It has to be small enough so that the user does not reject the system (e.g., 1\%).
\end{itemize}

Those two rates are tightly linked together: reducing the FRR for system convenience raises the possibility for an impostor to log into the smartphone. They are usually displayed as ROC curves (Receiver Operating Characteristic) as we will see in the result part.

\section{Similarity score computation}
Herein we describe three methods that allow to compute a similarity score between two images: the enrolled image and the candidate image, denoted as $E$ and $C$, respectively. These techniques come from the state of the art and could serve as a baseline in terms of recognition rate performance in the context of small fingerprint sensor. 

The first method uses all the pixels' intensity in a global manner to compute the score; it belongs to the class of direct methods \cite{Brown05,Szeliski10}. The two others use features and descriptors to perform matching between the two images; they are much more widespread than the former method nowadays \cite{Lowe04,Mikolajczyk05}.
We assume that the images are defined on a discrete regular grid of size $(m$x$n)$ pixels so that $E(x, y)$ denotes the pixel intensity of the image $E$ at pixel location $\mathbf{p}=[x, y]^T$.

\subsection{Direct approach}
The first method uses the zero mean normalized cross correlation (ZNCC) to compute a correlation surface $\gamma(u,v)$ between the two images by translating one with respect to the other:
\begin{equation}
\gamma(u,v) = \frac{\sum_{x,y \in \mathcal{S}}(E(x,y)-\mu^{E})(C(x-u,y-v)-\mu^{C})}{\sigma^{E}\sigma^{C}}
\end{equation}
for $u \in[-n+1,n-1]$ and $v \in[-m+1,m-1]$. The support $\mathcal{S}$ contains the set of pixels' coordinates $\{(x,y)\}$ so that $x-u\in[1,n]$ and $y-v\in[1,m]$; thus it depends on $u$ and $v$. Statistical parameters are computed each time for $u$ and $v$:
\begin{equation*}
\mu^{E} = \frac{\sum_{x,y \in \mathcal{S}} E(x,y)}{\text{Card}(\mathcal{S})} \text{ and }  
\mu^{C} = \frac{\sum_{x,y \in \mathcal{S}} C(x-u,y-v)}{\text{Card}(\mathcal{S})},
\end{equation*}
as well as:
\begin{align*}
\sigma^{E} &= \sqrt{\smash[b]{\textstyle \sum_{x,y \in \mathcal{S}} \big(E(x,y)-\mu^{E}\big)^2}},\\
\sigma^{C} &= \sqrt{\smash[b]{\textstyle \sum_{x,y \in \mathcal{S}} \big(C(x-u,y-v)-\mu^{C}\big)^2}}.
\end{align*}
The final similarity score is the maximum of the correlation surface:
\begin{equation}
s_{\text{cor}} = \max \gamma(u,v)
\end{equation}

There are at least two advantages to this method: it is robust with respect to any global illumination variation in the images and a fast implementation already exists \cite{Lewis95}. In practice, those two criteria have often given the advantage to this method compared to other more robust methods \cite{Chambon03}. However, this proposal is not invariant to rotations and its performance degrades with the amplitude of local warps between the two images. In order to cope with its sensibility to rotations, in this work, we compute several correlation scores between $E$ and $C$ rotated between -30 and 30 degrees with a step size of 5 degrees and we keep the final score.

\subsection{Feature based approach}
Descriptors have replaced direct approaches since many years in image registration problems and have extended their capabilities in other domains like pattern recognition. The used scheme is quite common between all their applications. We summarize here the relevant steps for our study:
\begin{enumerate}
\item \textbf{feature point extraction}: 
the objective here is to look for pixels' coordinates in both images $E$ and $C$, denoted as $\mathbf{p}^{E}$ and $\mathbf{p}^{C}$ respectively, such that the matching process is made easier after. The literature is rich on this subject \cite{Harris88,Lowe04,Mikolajczyk05}, but the minimum requirement for a feature point is that it should have contrast variation in its neighborhood; in other words spatial derivative of the image function around this point should not be zero. As a result of this step, we have two sets, $\{\mathbf{p}^{E}\}$ and $\{\mathbf{p}^{C}\}$, not necessarily of the same dimension.
\item \textbf{descriptors computation}: 
also based on the idea of matching, the descriptor tries to characterize the surrounding region of the key point. It has to be discriminant and compact in the quantity of information it caries so as to be efficient in terms of computational speed. Again, the literature is rich: e.g., methods using spatial repartition of points to record the shape \cite{Belongie02}, histogram of gradient distribution \cite{Lowe04}, oriented raw patches \cite{Brown05}; this list is far from being exhaustive, we refer to \cite{Szeliski10} for a comprehensive review of this topic.
At the end of this stage, each point $\mathbf{p}_i$ has been assigned a descriptor $\mathcal{D}_i$, which could take the form of a vector or a matrix containing various information (histogram, raw intensity patch, etc.).
\item \textbf{matching}:
this step searches for the links between each pair of points from the sets $\{\mathbf{p}^{E}\}$ and $\{\mathbf{p}^{C}\}$ by comparing the points' descriptors. The easiest strategy is to link each point $i$ of $E$ to the point in $C$ that minimizes the distance $h$ between their descriptors. The search is exhaustive:
\begin{equation}
\hat{k} = \underset{k}{\operatorname{argmin}} ~ h(\mathcal{D}^E_i, \mathcal{D}^C_k),
\label{eq_mec}
\end{equation}
where $\hat{k}$ is the index of the element of $\{\mathcal{D}^{C}\}$ chosen for the matching with $\mathcal{D}^E_i$. The function $h$ can be for example an euclidean or Hamming distance w.r.t. the kind of chosen descriptor. So as to get a bijection between the two sets of points, the same procedure can be realized in reversed order. As a consequence, only the matching results which satisfy the minimum cost criterion symmetrically are kept; this is the method used in this work. We finally obtain a set of linked points $\{\mathbf{\bar{p}}^{E},\mathbf{\bar{p}}^{C}\}$. Notice that more elaborate strategies exist, e.g., the Hungarian algorithm, which looks for an optimal coupling at a minimal cost in a bipartite graph \cite{Belongie02}, or methods that sort the descriptors into trees or hash table to speed up their matching \cite{Lowe04,Brown05}. It is also possible to increase the robustness with a ratio test of the descriptor \cite{Lowe04} with the aim to understand how much the correspondence in the list is better than the next one.

So as to satisfy a geometrical constrain in our problem, which is linked to the fact that the finger can only rotate and translate at the sensor surface in between two frames, an additional step of function minimization allows to exclude potential outlying matches:
\begin{equation}
\hat{\theta}, \hat{\mathbf{t}} = \underset{\theta, \mathbf{t}}{\operatorname{argmin}} \sum_i \rho\big(\mathbf{\bar{p}}^{E}_i - R(\theta) (\mathbf{\bar{p}}^{C}_i+\mathbf{t})\big)
\end{equation}
where $R(\theta)$ stands for the rotation matrix in the image plane and $\rho$ is a function allowing for the robust estimation of $\hat{\theta}$ and $\hat{\mathbf{t}}$. The optimized parameters are not of interest here, since they only serve the geometrical constraint to find a new set of points $\{\mathbf{\bar{\bar{p}}}^{E},\mathbf{\bar{\bar{p}}}^{C}\}$ which satisfy:
\begin{equation}
\|\mathbf{\bar{\bar{p}}}^{E}_i - R(\theta) (\mathbf{\bar{\bar{p}}}^{C}_i+\mathbf{t})\| < \epsilon,
\end{equation}
for all indices $i$ of elements in the set of points, where $\|.\|$ is the euclidean norm and $\epsilon$ a small positive scalar. This allows to define the similarity score between $E$ and $C$ that has been chosen for the feature based methods in this article:
\begin{equation}
s_{\text{desc}} = \text{Card}(\{\mathbf{\bar{\bar{p}}}^{E},\mathbf{\bar{\bar{p}}}^{C}\}).
\end{equation}
Basically, the score is the number of good matches between the two images after all the preceeding steps. The function $\rho$ can take multiple forms, like belonging to the class of the M-estimators. In practice, we use the RANSAC algorithm \cite{Fischler81} because of its good fit to this kind of problem.
\end{enumerate}

It should be noticed that the suggested matcher is not necessarily efficient in terms of computational speed; this is not the goal of our work, which is focused on the two first steps. We observed experimentally that RANSAC makes the process flow quite insensitive to the choice of a "smarter" matching procedure than the naive and exhaustive method we suggest. Yet it would be interesting to avoid RANSAC by using a better strategy for robustly pairing the descriptors. We will now describe the two feature based approaches considered in this paper.

\subsubsection{SIFT}
This algorithm, which is the acronym of Scale Invariant Feature Transform, has been very popular in the early 2000s \cite{Lowe04} and keeps being a key reference in computer vision. It optimizes all the steps previously described to achieve detection and matching of features (also named key points) that possess the following properties:
\begin{itemize}
\item robustness with respect to affine warps, scale and illumination invariance between to images (step 1),
\item optimized processing time thanks to compact and discriminant descriptors (step 2), and their sorting in trees to speed up the matching phase (step 3).
\end{itemize}

In this work, we chose to only use the SIFT key points detection together with their descriptors. The matching part follows the same scheme that we have previously described because this will facilitate the comparison with the second method. The consequences are a longer processing time with more outlying matches, but these are automatically corrected using the RANSAC algorithm. 

The SIFT features are selected across multiple scales as extrema of the Difference of Gaussians applied to the image. Intuitively, it would be like filtering the image to extract its edges at several scales and then extract coordinates of pixels which are local extrema along the edges and in between adjacent scales. Features with low contrast and weak curvature along the edge are discarded. The remaining ones are again processed to get a sub-pixel location refinement as well as an orientation value estimated with the surrounding image gradient. This completes step 1.

The SIFT descriptor tries to capture in histograms an orientation information of the gradients in a window size of (16x16) pixels around each key point. A weighting scheme of those histograms makes the descriptor robust to illumination changes and invariant to rotations. The SIFT descriptor is a concatenation of histograms values, and the distance function between two of them is the euclidean distance:
\begin{equation}
h(\mathcal{D}_i, \mathcal{D}_k) = \| \mathcal{D}_i-\mathcal{D}_k \|.
\end{equation}
This completes steps 2 and 3 for SIFT in our case.

\subsubsection{Harris-SSD}
This method combines the Harris corner detector \cite{Harris88} for feature point detection and the Sum Square Differences for descriptor comparison, hence its name.\\
Harris and Stephen's algorithm uses a Taylor expansion of the squared differences between an image and itself slightly translated. This expression can be written locally for each pixel in a matrix form:
\begin{equation}
H_{x,y} = \nabla_{\sigma_H} E(x,y) \nabla_{\sigma_H} E(x,y)^T,
\end{equation}
where $H_{x,y}$ is the Harris matrix of size (2x2) at pixel $(x,y)$ for image $E$ and $\nabla_{\sigma_H} E(x,y)$ the spatial derivative along $x$ and $y$ of $E$ smoothed with a Gaussian window of parameter $\sigma_H$  set to $1$ in our study. An eigenvalue analysis of $H_{x,y}$ tells if the pixel belongs to a uniform region, an edge, or a corner. This latter being able to resolve all spatial ambiguities, it becomes the best feature candidate after thresholding the Harris function on the image. Associated to each feature, we compute the local orientation $\theta$ using the gradient of the smoothed image in the feature neighborhood:
\begin{equation}
\theta(x,y) = \arctantwo\big(\nabla^y_{\sigma_{\theta}} E(x,y), \nabla^x_{\sigma_{\theta}} E(x,y)\big),
\end{equation}
where the smoothing parameter $\sigma_{\theta}$ is set to $4$. Therefore, each feature point has a location $\mathbf{p}$ and an orientation $\theta$ associated to it; this completes step 1.

The descriptor used here is an oriented image patch around the feature point; this patch is centered and normalized with respect to its pixel intensity values so as to reduce its sensitivity to lighting and skin conditions. Let $I_{\mathbf{p},\theta}$ be the patch of size (2$w$+1x2$w$+1) which is extracted around the feature location $\mathbf{p}$ with orientation $\theta$. We have:
\begin{equation}
\mathcal{D} = \frac{I_{\mathbf{p},\theta}-\mu^{I_{\mathbf{p},\theta}}}{\sigma^{I_{\mathbf{p},\theta}}}.
\end{equation}
The patch is extracted using interpolation relative to its orientation $\theta$ so that the strongest gradients are detected along the $x$ axis of $I_{\mathbf{p},\theta}$. The window size parameter $w$ is set to 7 in this work. Consequently, $\mathcal{D}$ is a matrix and the distance function between two chosen descriptors is:
\begin{equation}
h(\mathcal{D}_i, \mathcal{D}_k) = \sum_{x,y} \big(\mathcal{D}_i(x, y)-\mathcal{D}_k(x, y)\big)^2.
\end{equation}
The function $h$ is also known as the SSD error criterion. This finalizes step 2 and 3 for the Harris-SSD method. The latter is actually a simplified version of the MOPS method (Multi Scale Oriented Patches) presented in \cite{Brown05} and successfully used for image mosaicking applications.

\section{Evaluation protocol}
This section describes the processed dataset for comparing the three methods as well as the evaluation criterion. Algorithms were implemented in Python and make extensive use of the scikit-image toolkit \cite{scikit-image}.

\subsection{Dataset}
We used the access free SDUMLA-HMT biometric dataset \cite{Yin11}. It contains fingerprint images of 106 persons acquired with five different sensors. Six fingers of each individual were captured eight times: the thumb, the forefinger and the middle finger for both hands. We focused our experimentations on the data recorded with the capacitive sensor FT-2BU.\footnote{Miaxis Biometrics sensor at 350dpi.} From this database, we randomly selected 30 people, corresponding to 1440 images of size (152x200) pixels. The data are labeled because all acquisitions are classified into multiple directories per individual and per finger. However, since the images of a same finger are not registered there could be translation and rotation in between acquisitions. Also, no enhancement filter is used in this work to compensate for artifacts such as a dry skin. We processed the raw images directly.

\subsection{Small sensor simulation}
For each fingerprint image, we cropped a patch of (70x70) pixels to simulate the small sensor acquisition. This size corresponds to a (5x5)mm${}^2$ imaging area for a 350dpi sensor, which is close to a smartphone sensor but at a lower resolution.
\begin{figure}
\center
\includegraphics[scale=0.5]{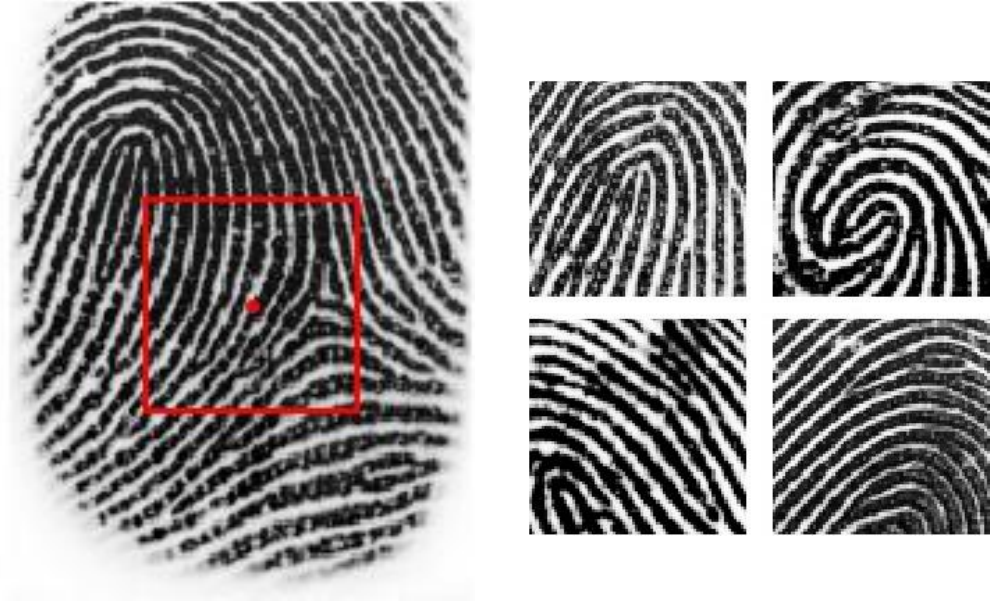}
\caption{Patch extraction on an acquisition (red bounds), and sample examples.}
\label{fig_processus_extraction}
\end{figure}

To insure that the extracted patch is centered with respect to the fingerprint, we used a k-means algorithm to segment the background in the original acquisition. The center of the patch is assumed to be the center of the bounding box around the fingerprint; this is illustrated in figure \ref{fig_processus_extraction}.

\subsection{Data clustering}
The patch images are then divided into two sets: the enrollment set which would correspond to the data recorded by the smartphone user during the initialization of the biometric system, and the candidate set which would contain genuine and impostor attempts captured by the sensor during its usage. For each finger, we selected a random set of $n$ enrollment images, the remaining ones being automatically classified as candidates.

\subsection{Evaluation criterion}
We computed score tables by making an exhaustive comparison between each candidate image and each enrollment image; this gives a score for each couple \{candidate image, enrolled finger\} which is a function of all the individual scores of the candidate image with each image of the enrolled finger. In the following results we employed the maximum of all of the scores. The score table allows to compute FRR and FAR at several thresholds. All three algorithms are compared using the resulting ROC curves from the dataset.

\section{Results}
\begin{figure}
\center
\includegraphics[scale=0.3]{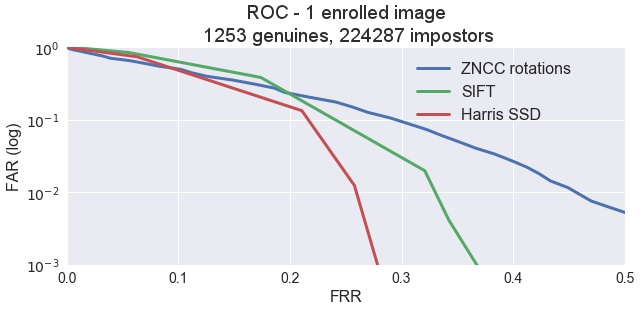}\\
\includegraphics[scale=0.3]{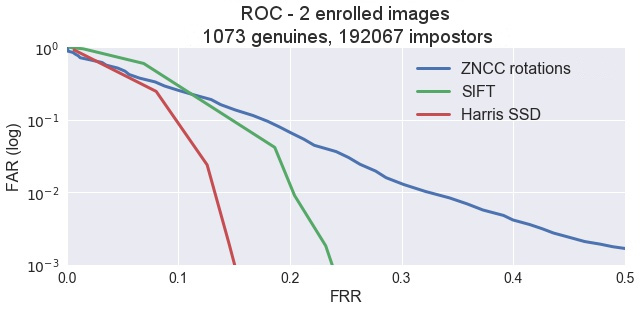}\\
\includegraphics[scale=0.3]{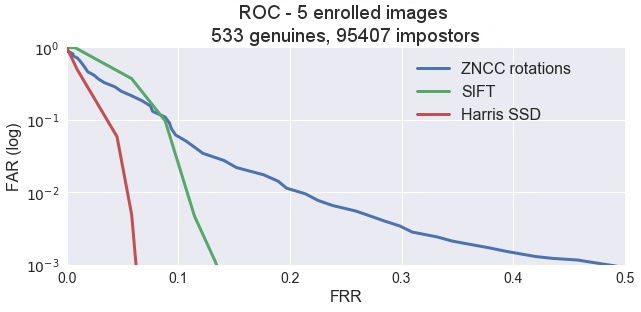}\\
\caption{ROC curves for the three methods with respect to the enrollment process.}
\vspace{-0.3cm}
\label{fig_roc}
\end{figure}

Figure \ref{fig_roc} summarizes the authentication performance of the three methods with respect to the number of images used for the enrollment process. The closer the ROC curve is to the point (0,$10^{-3}$), the better the algorithm performs.
We observe that increasing the number of enrolled images drastically reduces the FRR at a fixed FAR for feature based methods. This is also the case for the ZNCC but to a lesser extent. We considered only five images at most for the enrollment process because of the limited size of the dataset, but ten images is a minimum during a real enrollment in a smartphone. Increasing the enrollment data size appears to be a good way to improve performance of any biometric system.

\begin{figure}
\center
\includegraphics[scale=0.43]{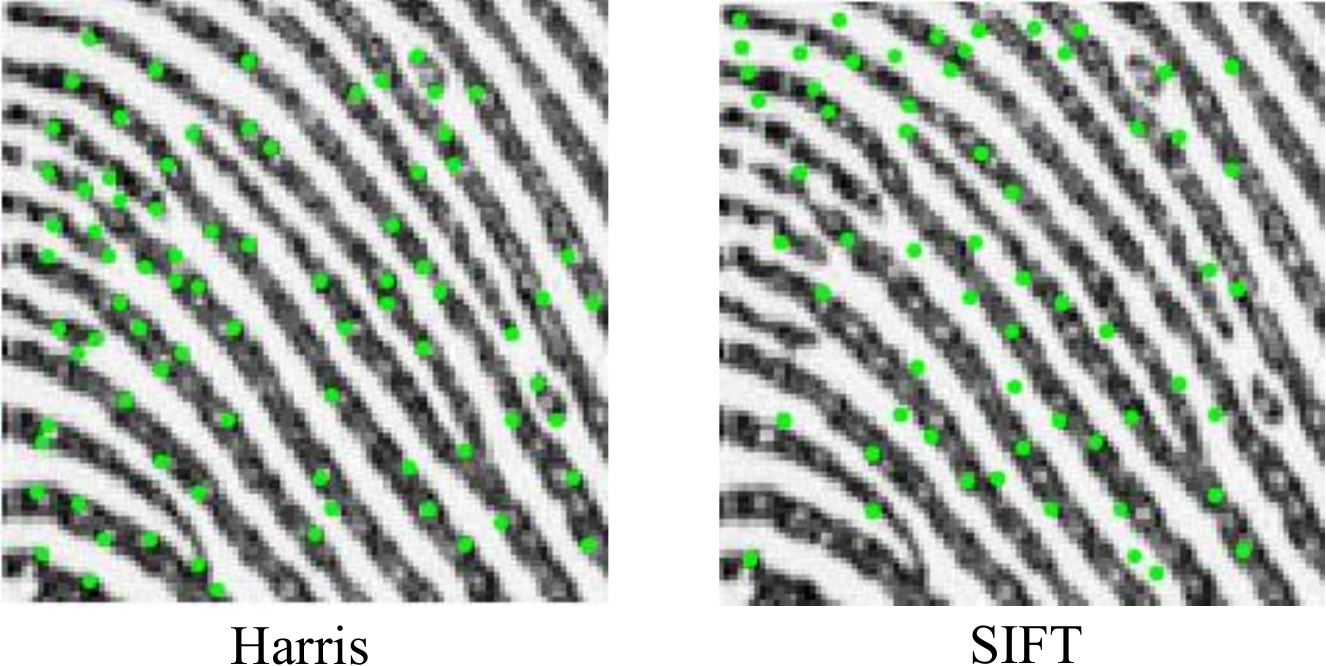}
\caption{Harris and SIFT features in green dots.}
\label{fig_points}
\vspace{-0.3cm}
\end{figure}
The Harris-SSD method also quite significantly outperforms SIFT. This observation contradicts several results in the literature like in \cite{Karlsson16}. It could be explained by the nature of SIFT which, as a generic feature transform, has many degrees of freedom like scale invariance. On the contrary, the Harris SSD method has been tailored to the geometrical constraints of our problem. However, we believe that the low performance of SIFT in this case comes from its feature detection step as it seeks extrema along ridges and valleys of the fingerprint where a lot of ambiguities could appear. This observation is enforced by figure \ref{fig_points} which provides a comparison of both detectors, SIFT and Harris corner, on the same image. Interestingly, the Harris corner is able to detect a lot of features like minutiae, sweat pore and discontinuities of a ridge, whereas SIFT is stuck along the valleys and ridges, which may be less discriminative regions. Finally, figure \ref{fig_points} displays some of the patches extracted around the Harris corners. The quantity of information they carry is difficult to reduce to a smaller descriptor; this was one motivation for keeping the raw image patch in the Harris-SSD method.
\begin{figure}
\center
\includegraphics[scale=0.27]{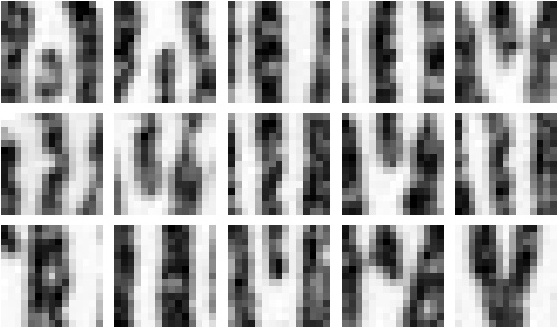}
\caption{Examples of descriptors (15x15 pixels) around Harris corners of figure \ref{fig_points}.}
\label{fig_patchs}
\vspace{-0.3cm}
\end{figure}

\section{Perspectives}
From our experience, there are at least three simple ways to improve the results in figure \ref{fig_roc}: 
\begin{itemize}
\item increase the number of images in the enrollment process; most of the failure cases in our results came from a lack of overlap between candidate and enrolled images.
\item Prefilter the images to compensate for skin and sensor artefacts; this is mandatory for any biometric system, especially the minutiae based methods. Yet we were quite surprised by the robustness of the Harris-SSD method in processing raw images directly.
\item Have a greater acquisition resolution; we displayed results using a 350dpi sensor while smartphones have 500dpi sensors. A better resolution will improve the features extraction as well as the discriminative property of the descriptor.
\end{itemize}
On a research perspective, it seems like the enrollment process still lack some clear rules for its validation. Yet it does impact the performance of the biometric system.

{\small
\bibliographystyle{ieee}
\bibliography{egbib}
}

\end{document}